\pdfoutput=1
\documentclass[10pt,letterpaper]{article}
\usepackage{cogsci}
\usepackage{pslatex}
\usepackage{apacite}
\usepackage{comment}
\usepackage{graphicx}
\usepackage{graphbox}
\usepackage{nicefrac}
\usepackage{lipsum}
\usepackage{amsmath,amssymb}
\usepackage[hyphens]{url}

\title{Emulating Human Developmental Stages with Bayesian Neural Networks}

\author{{\large \bf Marcel Binz (binz@staff.uni-marburg.de)} \\
  Department of Psychology, Theoretical Neuroscience Group \\
  Philipps-Universit\"at Marburg
  \AND {\large \bf Dominik Endres (dominik.endres@staff.uni-marburg.de)} \\
  Department of Psychology, Theoretical Neuroscience Group \\
  Philipps-Universit\"at Marburg}

\begin{document}

\maketitle

\begin{abstract}
We compare the acquisition of knowledge in humans and machines. Research from the field of developmental psychology indicates, that human-employed hypothesis are initially guided by simple rules, before evolving into more complex theories. This observation is shared across many tasks and domains. We investigate whether stages of development in artificial learning systems are based on the same characteristics. We operationalize developmental stages as the size of the data-set, on which the artificial system is trained. For our analysis we look at the developmental progress of Bayesian Neural Networks on three different data-sets, including occlusion, support and quantity comparison tasks. We compare the results with prior research from developmental psychology and find agreement between the family of optimized models and pattern of development observed in infants and children on all three tasks, indicating common principles for the acquisition of knowledge. 

\textbf{Keywords:} 
Core knowledge; developmental psychology; intuitive physics; approximate number system; machine learning, deep learning, variational inference, normative models
\end{abstract}

\section{Introduction}

The theory of core knowledge in developmental psychology identifies several domains, that build the foundations of human cognition \cite{spelke2007core, lake2017building}. Typically physics, actions, numbers, space and social interactions are listed among the core domains. Knowledge in these areas is present starting from early stages of childhood and serves as the basis for learning during later life. Research in developmental psychology over the past decades equipped us with a solid understanding about the acquisition of such knowledge. Different stages of development have been identified for a wide range of phenomena. Insights across studies suggest, that established rules generally start with a simple hypothesis before becoming more sophisticated over time \cite{baillargeon2002acquisition}. \\

We investigate whether current machine learning systems show generalization behavior reminiscent of human infants and children at different stages of development. For this purpose, we assume that the amount of data available to the learning algorithm is proportional to human age. The class of models we focus on are Bayesian Neural Networks (BNNs), that are trained through variational inference. Neural networks have the desirable property of being able to approximate an arbitrary complex mapping given enough capacity, while Bayesian inference captures normative principles of how to update an initial belief in the light of new evidence. The specific choice of variational inference and neural networks is mainly due to convenience reasons, and we hypothesize, that many different combinations of universal function approximators and Bayesian learning would lead to comparable results. \\

Our experiments focus on two of the established core domains: physics and numbers. We consider two experiments involving intuitive reasoning about the laws of physics \cite{baillargeon2002acquisition} and one examining the approximate number system, which is responsible for forming fast, but imprecise, representations of quantities \cite{halberda2008developmental}. In all three cases we observe pattern in our models, that share similarities with the development progress during childhood, as we increase the data-set size. \\

There has been a recent interest in replicating reasoning capabilities from the core domains in artificial systems. Prior work regarding intuitive physics has considered generative \cite{battaglia2013simulation, chang2016compositional} as well as discriminative models \cite{lerer2016learning}. Both classes of models are often able to reach performance levels comparable to those of adults on specific tasks. Another core domain, that has received some attention within the computational modelling community, is the one of intuitive psychology. Here, for example, \citeA{baker2009action} suggest to employ Bayesian inverse planning for inferring mental states of other agents. In contrast to the aforementioned prior work, we are interested in the differences between optimal models for varying data-set sizes and how these differences compare to observations made in developmental psychology. Existing work on modelling the \emph{development} of intuitive physics is limited to descriptive models, such as list of rules \cite{siegler1998developmental} or decision trees \cite{baillargeon2009account}. In contrast to this, our approach is based on normative principles and we ask the question, whether observed stages emerge naturally in complex artificial learning systems. \\

In the next section we provide a short technical overview of neural networks and variational inference. This is followed by a description of the three experiments under examination. For each experiment we outline the given task, the empirical observations made in the developmental psychology literature and how we construct an artificial data-set. We finally provide a comparison between the developmental progress of children at different ages and that of optimized models for different data-set sizes. We conclude the article with a discussion of the obtained results and an outlook of the future interaction between the areas of machine learning and developmental psychology.

\section{Methods}

\subsection{Deep Learning}

Neural networks are parametric function approximators, that combine linear transformations $\mathbf{W}$ and non-linear activation functions $f$ in alternating fashion: 
\begin{equation*}
    \mathbf{h}_l = f_l \left( \mathbf{W}_l^{\top} \mathbf{h}_{l-1} \right)
\end{equation*}
where $l \in \{1, \ldots, L\}$. $\mathbf{h}_0$ corresponds to the input $\mathbf{x}$ and $\mathbf{h}_L$ to an estimate of the target $\hat{y}$. Parameters of the model are commonly updated via gradient descent on a loss function, usually some form of negative log-likelihood. The power of neural networks stems from their ability to approximate any continuous function on a compact subset of $\mathbb{R}^n$.

\subsection{Variational Inference}

The task of learning model parameters $\mathbf{W}$ can also be stated as a Bayesian inference problem:
\begin{equation} \label{eq:Bayes}
    \underbrace{p(\mathbf{W} | \mathcal{D})}_{\text{posterior}} = \frac{\overbrace{p(\mathbf{y} | \mathbf{X}, \mathbf{W})}^{\text{likelihood}} \overbrace{p(\mathbf{W})}^{\text{prior}}}{\underbrace{p(\mathbf{y} | \mathbf{X})}_{\text{evidence}}}
\end{equation}

for a given data set $\mathcal{D} = \{(\mathbf{x}_i, y_i) \}_{i=1}^N$, with inputs $\mathbf{X} = \{ \mathbf{x}_i\}_{i=1}^N$ and targets $\mathbf{y} = \{ y_i \}_{i=1}^N$. Bayes' theorem defines how we should update our beliefs as more information becomes available. As $N \rightarrow \infty$ the influence of the prior vanishes, while for $N \rightarrow 0$ we can only rely on prior assumptions. In our context we assume that experience (i.e. $N$) increases with age and hence we use an approximation to Equation \ref{eq:Bayes} with data-sets of varying size to represent agents of different age. \\

Equation \ref{eq:Bayes} is in general hard to compute for models of useful complexity. Variational inference offers a tractable approximation to Equation \ref{eq:Bayes} \cite{hinton1993keeping}. Let $q_{\phi}(\mathbf{W})$ be a distribution with parameters $\phi$, that approximates the true posterior $p(\mathbf{W} | \mathcal{D})$. Formulating the problem as a minimization of the Kullback-Leibler (KL) divergence between $q_{\phi}(\mathbf{W})$ and  $p(\mathbf{W} | \mathcal{D})$ leads to the evidence lower bound (ELBO):
\begin{equation} \label{eq:elbo}
    \mathcal{L}(\phi) = \mathbb{E}_{q_{\phi}(\mathbf{W})} \left[ \log p(\mathbf{y} | \mathbf{X}, \mathbf{W})\right] - \text{KL}(q_{\phi}(\mathbf{W}) || p(\mathbf{W}))
\end{equation}

which can be maximized with respect to $\phi$ using standard optimization techniques. \\

In order to be able to scale to large data-sets Equation \ref{eq:elbo} is often approximated using batches $\mathcal{B} \subseteq \mathcal{D}$ of size M, with the log-likelihood term being scaled appropriately:
\begin{equation} \label{eq:N}
    \log p(\mathbf{y} | \mathbf{X}, \mathbf{W}) \approx \frac{N}{M}  \sum_{i \in \mathcal{B}} \log p(y_i | X_i, \mathbf{W})
\end{equation}

Note, that only the first term of the ELBO depends on the data-set size $N$, while the second term is independent of it. Hence the divergence term will dominate for small data-sets, leading to models that closely reflect our prior assumptions. In this work we employ priors, that promote simple functions. Therefore, our models are able to capture successively more complex functions with increasing data-set size.

\subsection{Implementational Details}

All models consist of $L = 3$ layers with hidden layer sizes $|h_l| = 256$ and ELU activation functions \cite{clevert2015fast}, unless otherwise mentioned. Inputs $\mathbf{x}$ correspond to flattened images of the scene and targets $y$ are dependent on the current task. We place a group horseshoe prior, which can be viewed as a continuous relaxation of a spike-and-slab prior \cite{mitchell1988bayesian}, over all parameters:
\begin{align*}
    &s \sim \mathcal{C}^+(0, \tau_0); ~~~ \tilde{z}_i \sim \mathcal{C}^+(0, 1); \\ 
    &\tilde{w}_{ij} \sim \mathcal{N}(0, 1); ~~~ w_{ij} = \tilde{w}_{ij}\tilde{z}_is
\end{align*}

and represent the approximate posterior $q_{\phi}(\mathbf{W})$ through a fully factorized distribution as proposed in \cite{louizos2017bayesian}. The sparsity hyperparameter of the horseshoe prior is fixed to $\tau_0 = 10^{-5}$. During training we approximate the expectation of the log-likelihood term with a single sample from $q_{\phi}(\mathbf{W})$ and make use of the local reparametrization trick \cite{kingma2015variational}. Gradient-based optimization is performed using Adam \cite{kingma2014adam} with batches consisting of $64$ samples. Results reported after training correspond to a Monte-Carlo estimate using 100 samples from $q_{\phi}(\mathbf{W})$.

\section{Experiments}

In this section we present an analysis of the proposed model on three different tasks adopted from the developmental psychology literature. The first two tasks involve reasoning about physical events (occlusion and support), while the last is concerned with the intuitive representation of quantities. For each task we include a summary of empirical observations made in children, alongside a comparison between these results and our models. Code for performing all experiments and generating artificial data-sets is publicly available\footnote{\url{https://github.com/marcelbinz/Developmental-Stages-of-BNNs}}.

\subsection{Occlusion Events}

\begin{figure*}[t]
    \centering
    \begin{minipage}{.29\textwidth}
    \centering
    \textbf{Original Task}
    \end{minipage}
    \begin{minipage}{.25\textwidth}
    \centering
    \textbf{Artificial Data}
    \end{minipage}
    \begin{minipage}{.42\textwidth}
    \centering
    \textbf{Results}
    \end{minipage} \\
    \begin{minipage}{.29\textwidth}
    \centering
    \includegraphics[width=\textwidth]{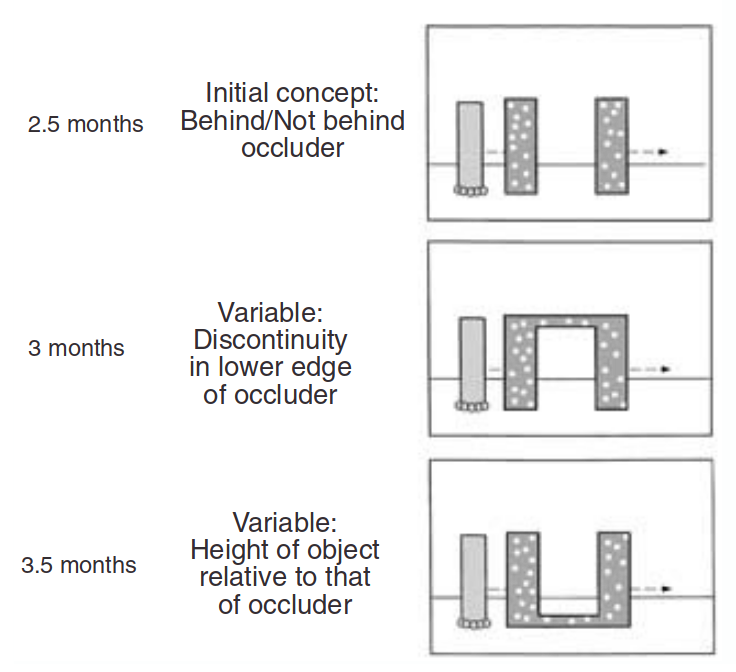}
    \end{minipage}
    \begin{minipage}{.25\textwidth}
    \centering
    \includegraphics[width=0.45\textwidth]{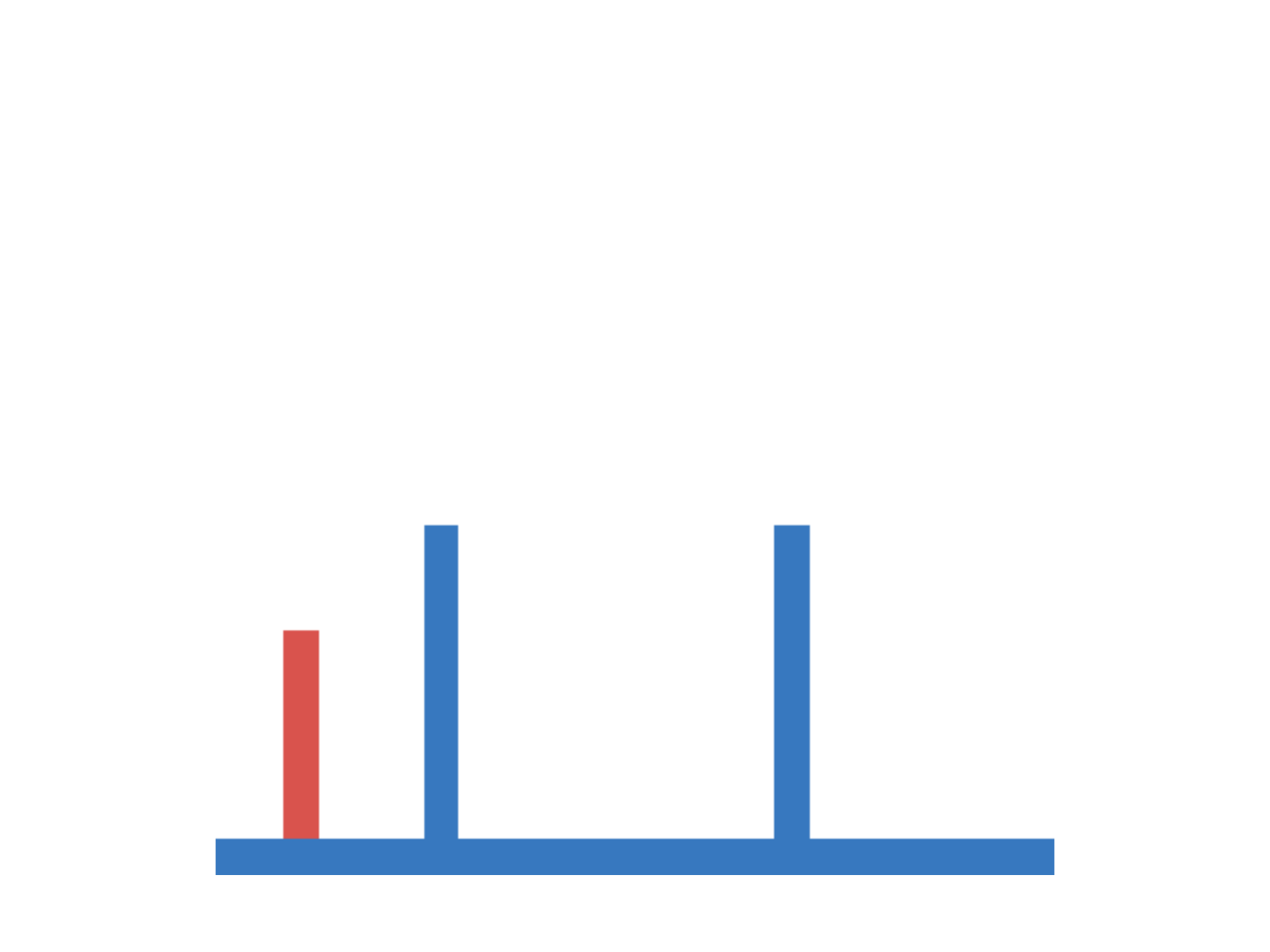} \\
    \includegraphics[width=0.45\textwidth]{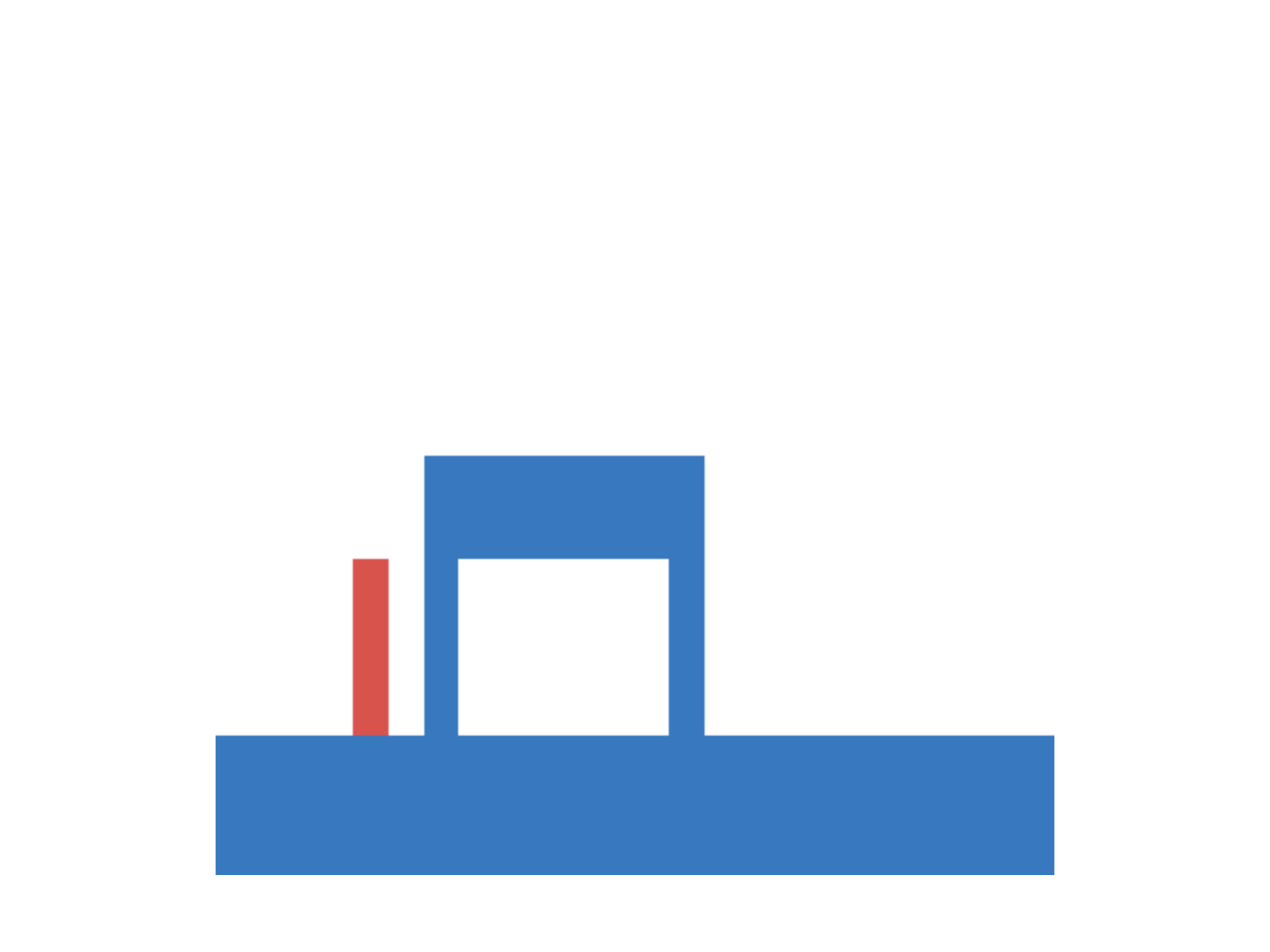} \\
    \includegraphics[width=0.45\textwidth]{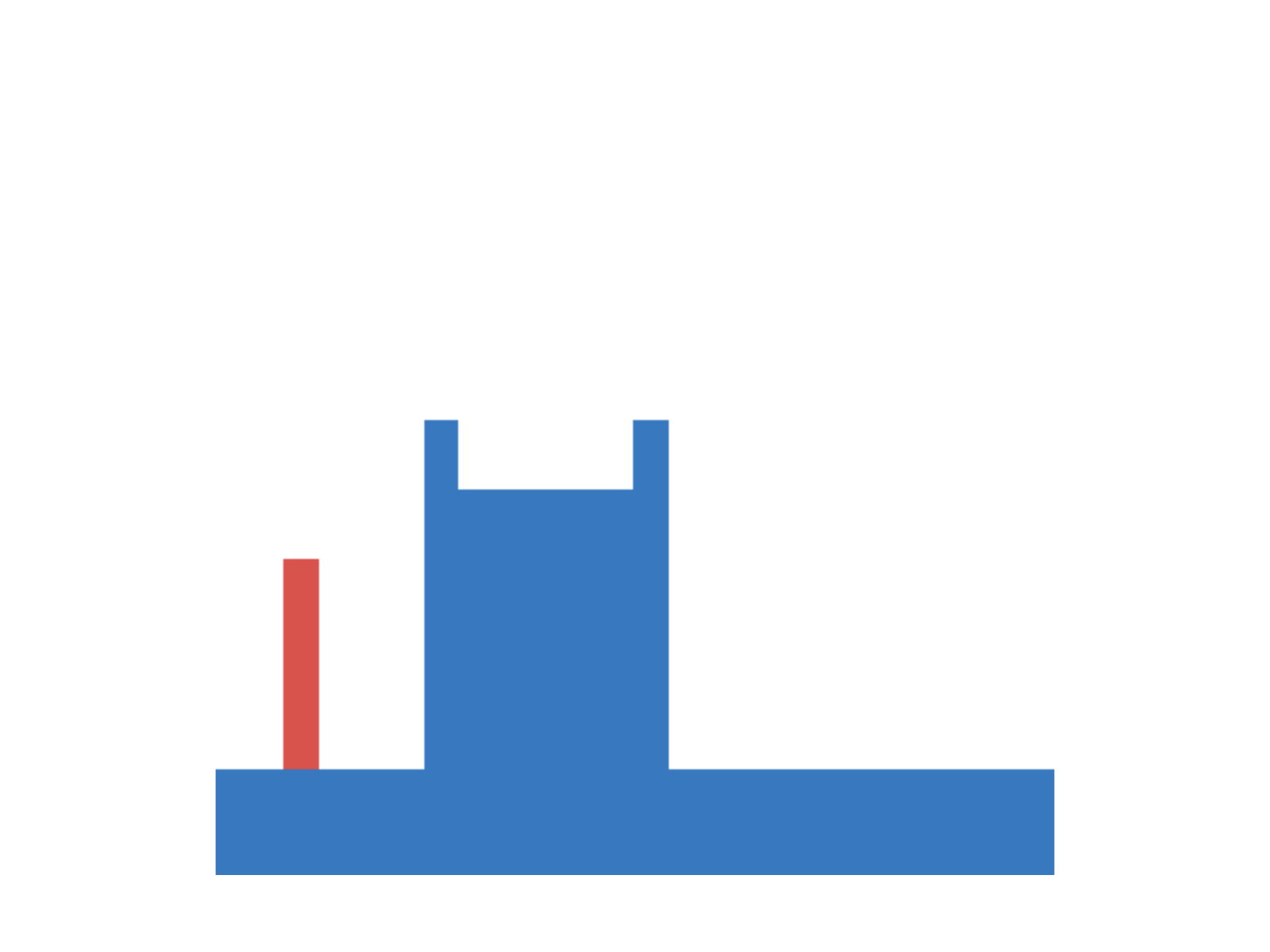} \includegraphics[width=0.45\textwidth]{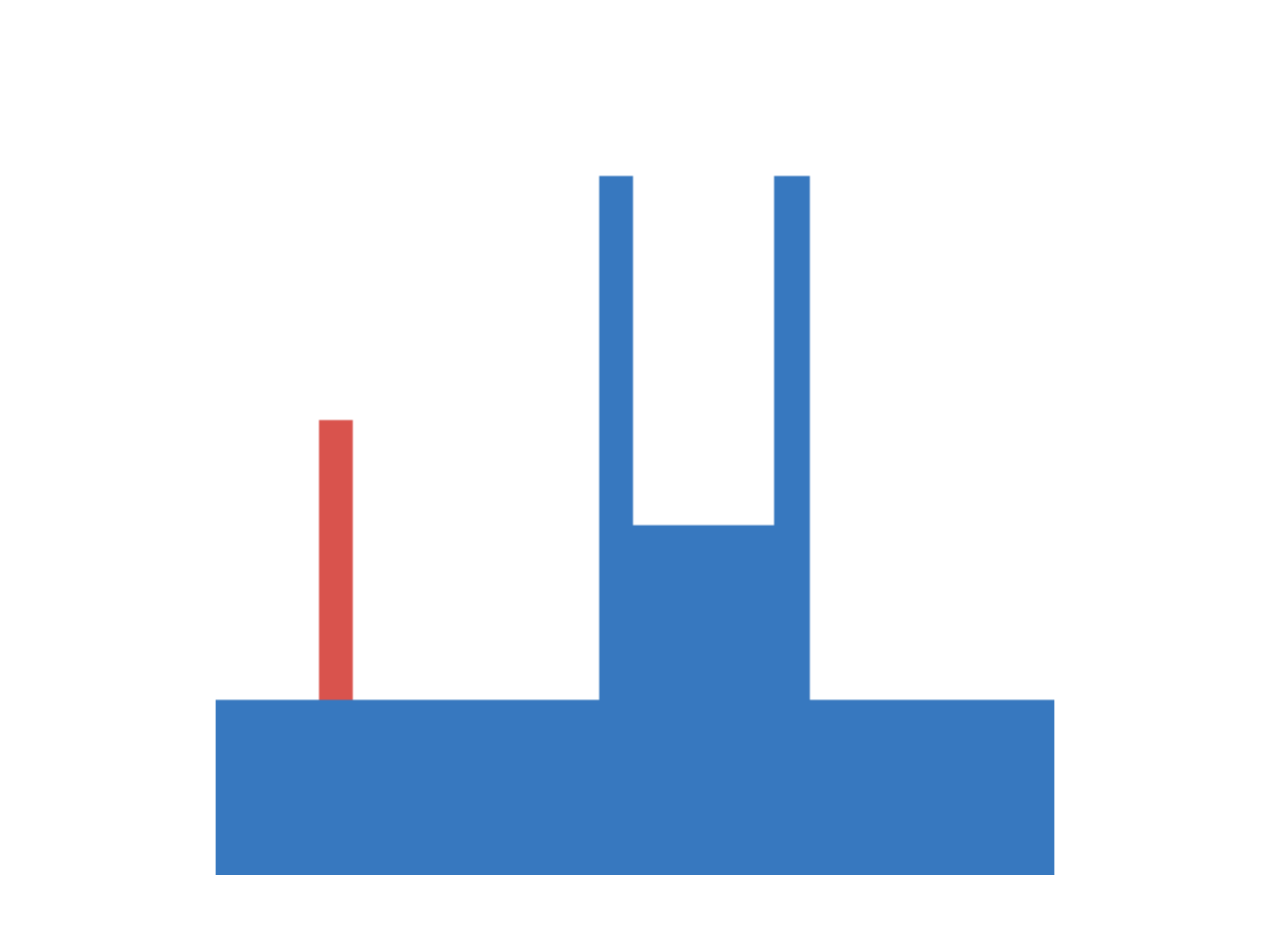}
    \end{minipage}
    \begin{minipage}{.42\textwidth}
    \centering
    \includegraphics[width=0.9\textwidth]{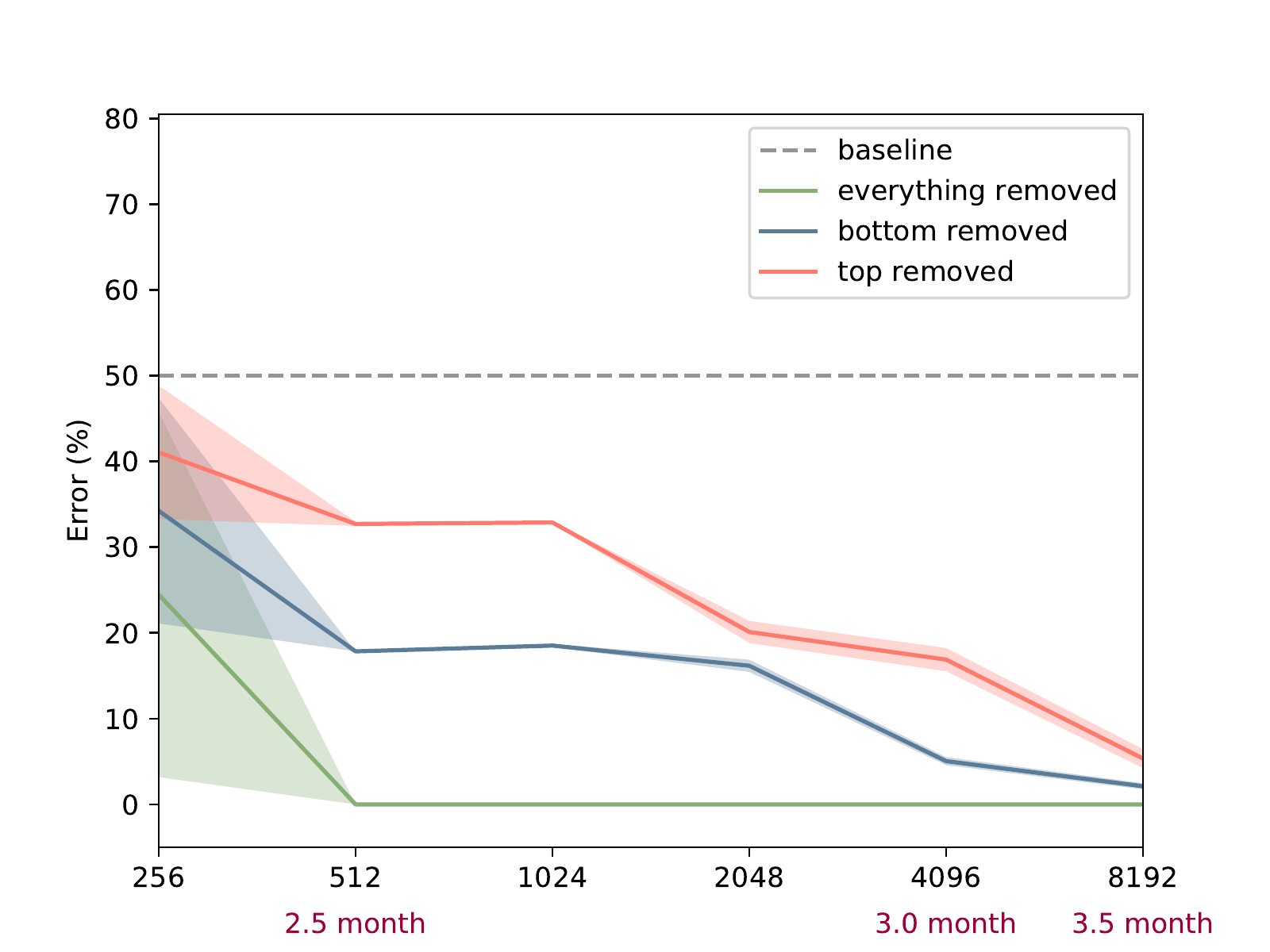}
    \end{minipage}
    
    \caption{Visualization of the occlusion event experiment. \textbf{Left:} Schematic illustration of the original setup. Figure adopted from \cite{baillargeon2002acquisition}. \textbf{Center:} Examples from the artificial data-set. \textbf{Right:} Performance for models with access to different amounts of data in the different conditions. Baseline indicates the chance for guessing randomly. Models discover solutions in order of their difficulty (green $\rightarrow$ blue $\rightarrow$ red), which is in accordance with observations made in the developmental psychology literature. }
    \label{fig:occu}
\end{figure*}

The first task under investigation is concerned with occlusion events. It is based on an experiment conducted by \citeA{baillargeon2002acquisition}. Each scene consists of a cylinder and a screen in form of a rectangular plane. During experimental manipulation the cylinder is moved back and forth behind the screen, while parts of the screen are removed. There are three different experimental conditions, each differing in which part of the screen is removed (top, bottom or everything removed). A depiction of the setup is shown in Figure \ref{fig:occu} (left). We are interested in infants' ability to judge, whether the cylinder remains visible as it moves behind the screen, which is measured via violation-of-expectation methods \cite{baillargeon1985object}. In violation-of-expectation methods gazing times for physically implausible events are measured. High gazing times for such events indicate, that children are surprised by the observation, which is interpreted as a violation of their expectation of what should have happened.  \\

Empirical evidence indicates, that infants form an initial concept based on a behind/not-behind distinction starting from the age of 2.5 months. At this stage they have learned, that the cylinder remains visible while moving past the screen, if the entire middle section of the screen is removed. They do not expect the cylinder to be visible in the other two conditions. This initial concept is refined during later stages of development. At the age of three months they also predict to see parts of the cylinder, if only the bottom part of the screen is removed. The knowledge of 3.5 months olds additionally extends to screens with removed top fractions. Note, that the last condition is more challenging compared to the other two, as it involves a comparison of heights between the cylinder and the lower connection of the screen (if the connection height is lower than the cylinder, the cylinder will be visible, otherwise it will not be visible). Baillargeon \emph{et al.} conclude from these observations, that infants start with initially simple rules about the laws of physics (behind/not-behind distinction), which become successively more sophisticated over time (reasoning about relative heights). \\

\begin{figure*}[t]
    \centering
    \begin{minipage}{.25\textwidth}
    \centering
    \textbf{Original Task}
    \end{minipage}
    \begin{minipage}{.25\textwidth}
    \centering
    \textbf{Artificial Data}
    \end{minipage}
    \begin{minipage}{.05\textwidth}
    \end{minipage}
    \begin{minipage}{.43\textwidth}
    \centering
    \textbf{Results}
    \end{minipage} \\
    \begin{minipage}{.25\textwidth}
    \centering
    \includegraphics[width=1.05\textwidth]{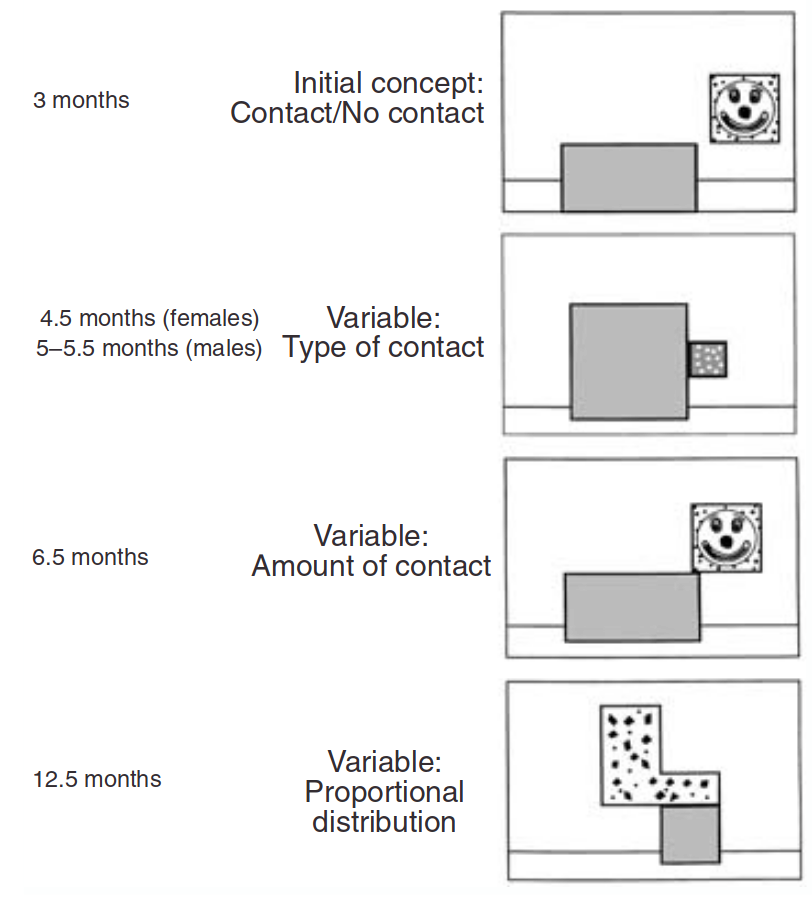}
    \end{minipage}
    \begin{minipage}{.25\textwidth}
    \centering
    \vspace{-0.18cm}
    \includegraphics[width=0.39\textwidth]{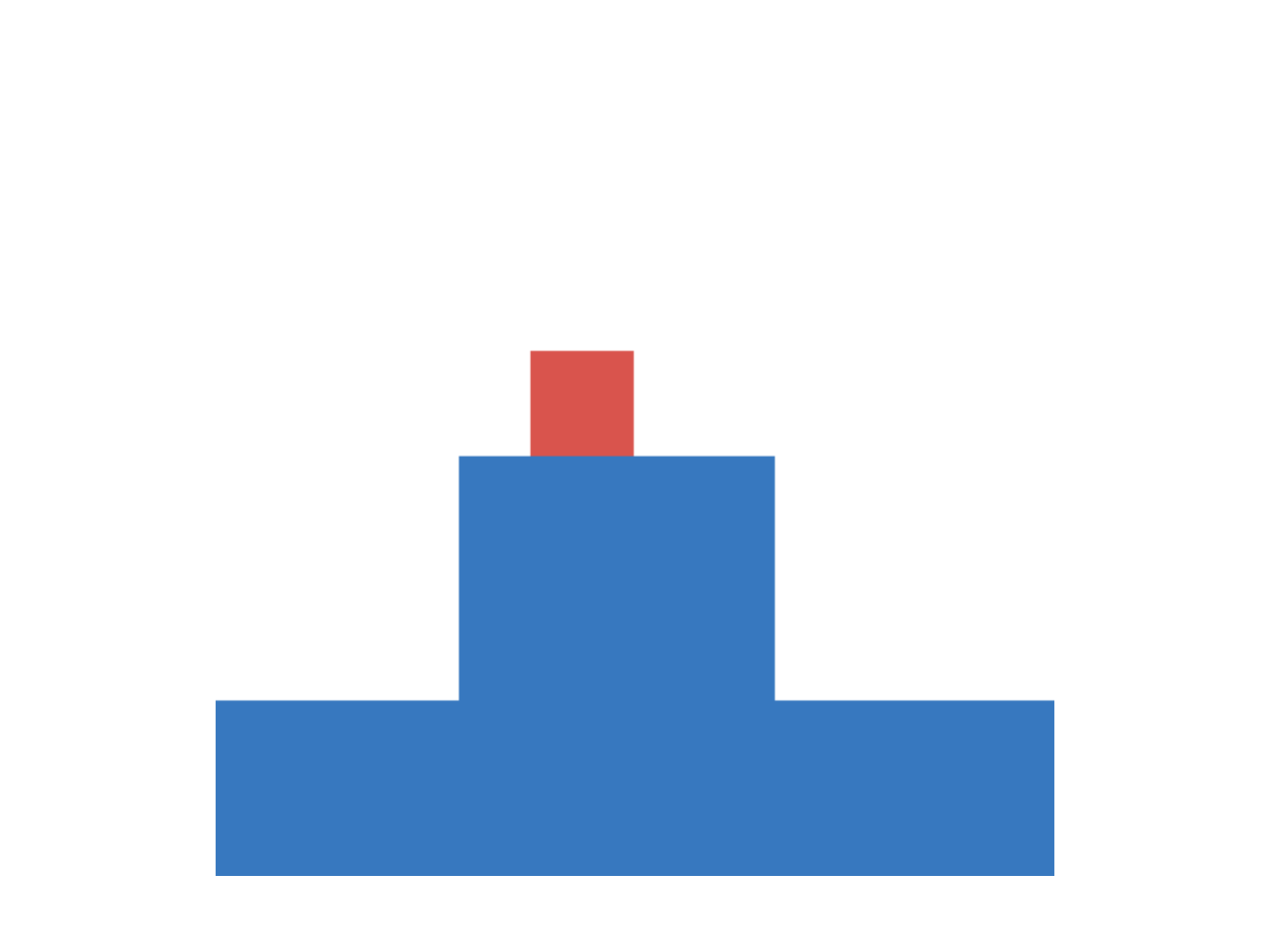} \includegraphics[width=0.39\textwidth]{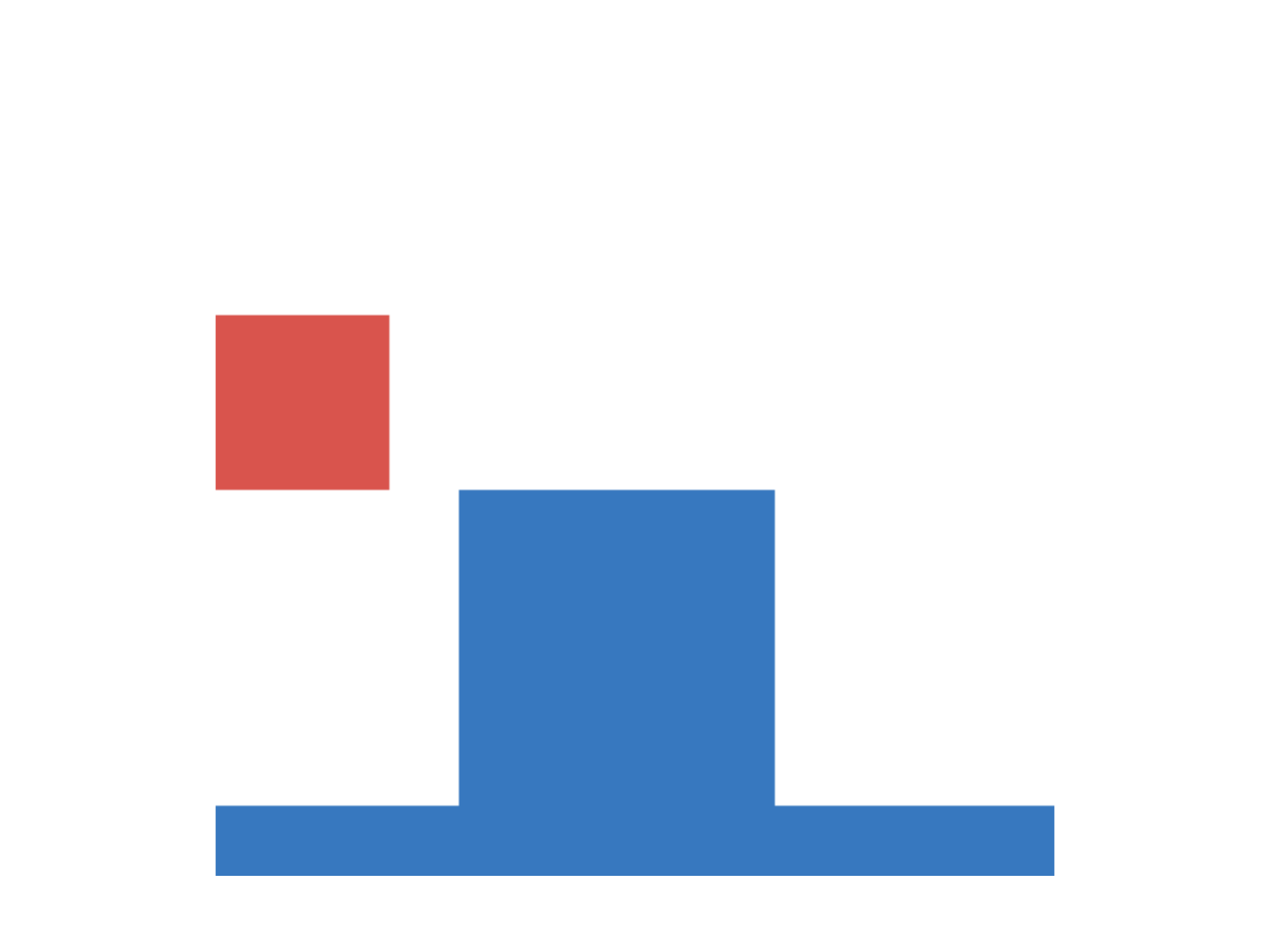} \\
    \includegraphics[width=0.39\textwidth]{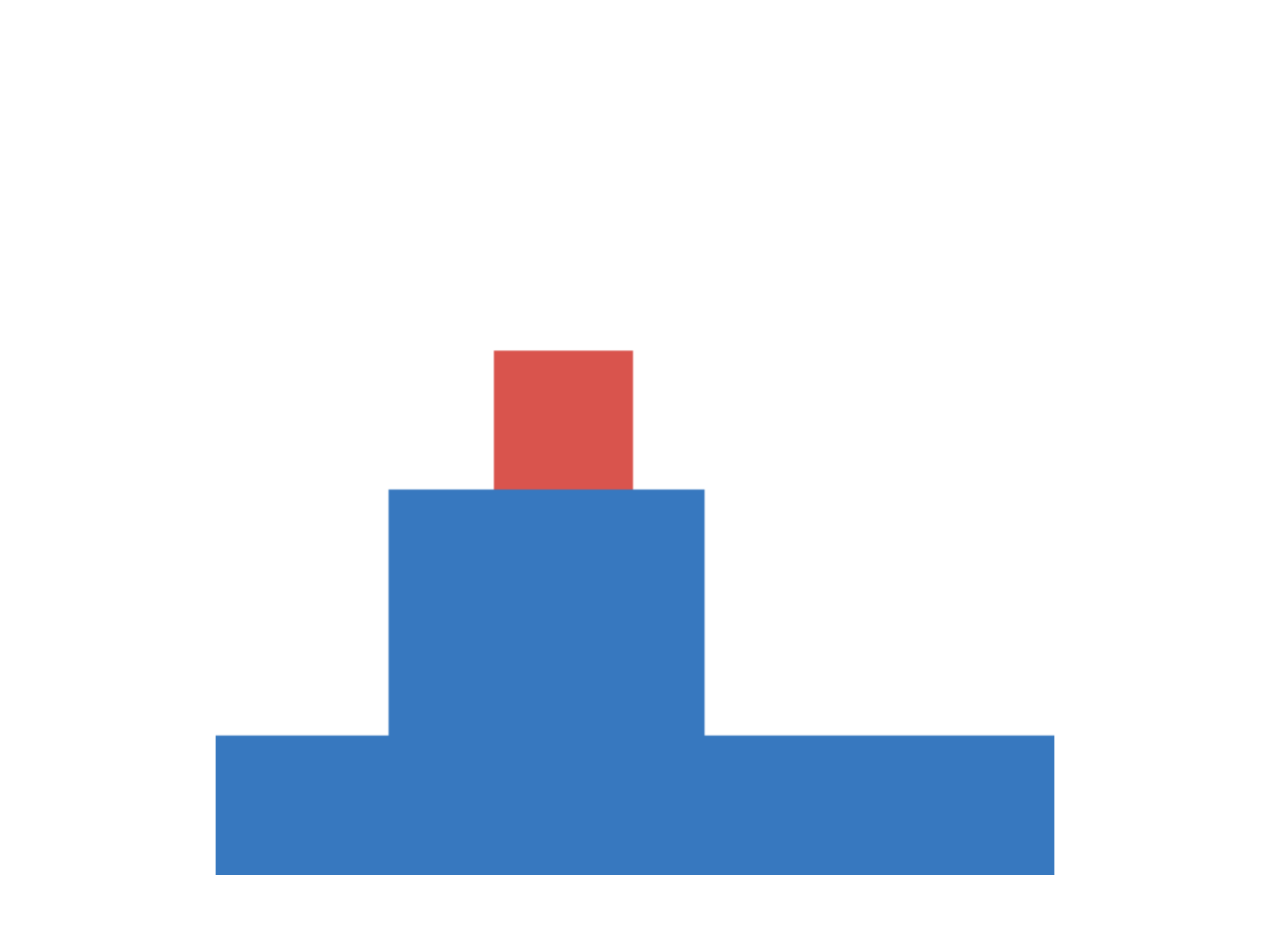} \includegraphics[width=0.39\textwidth]{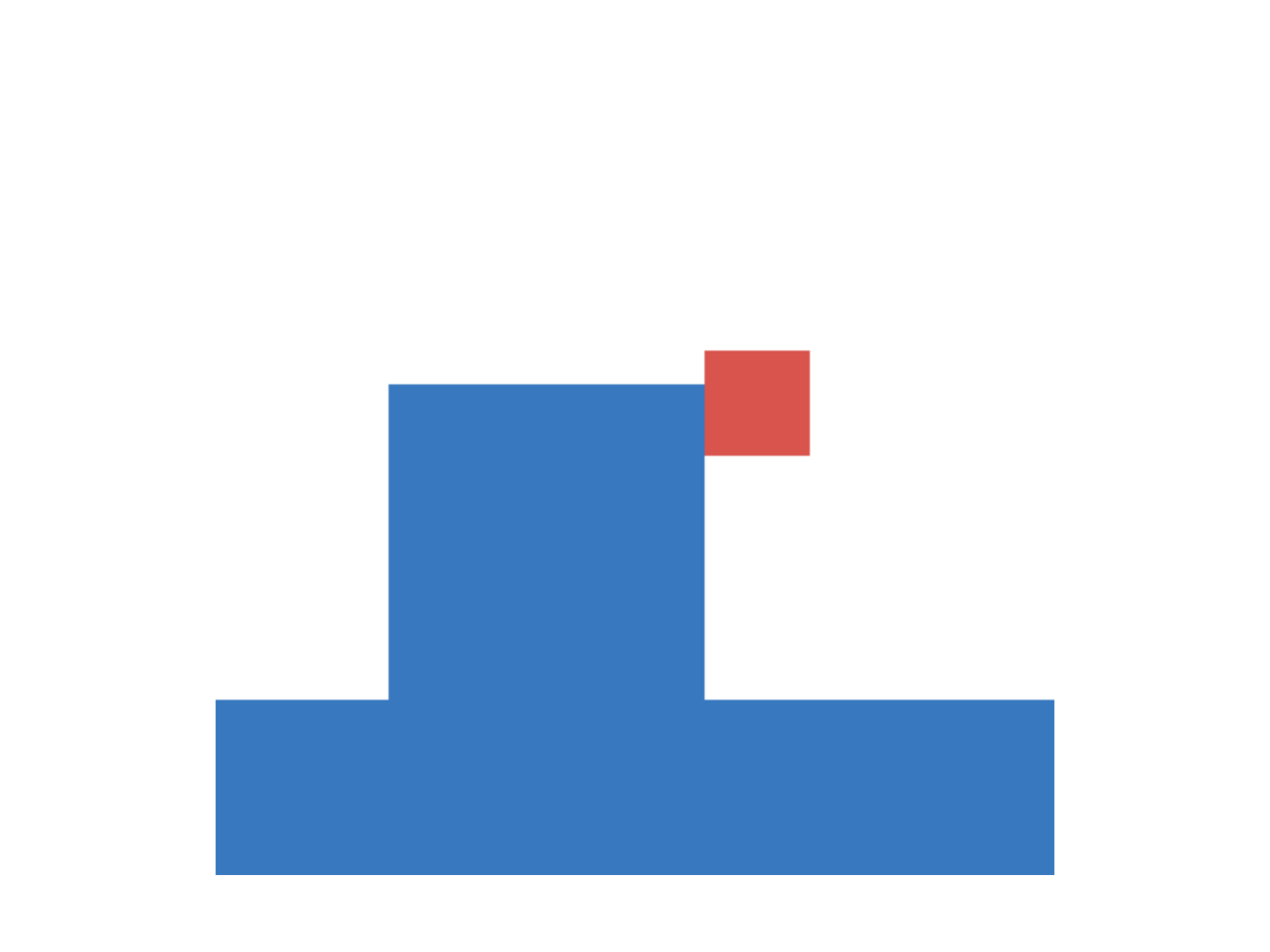} \\
    \includegraphics[width=0.39\textwidth]{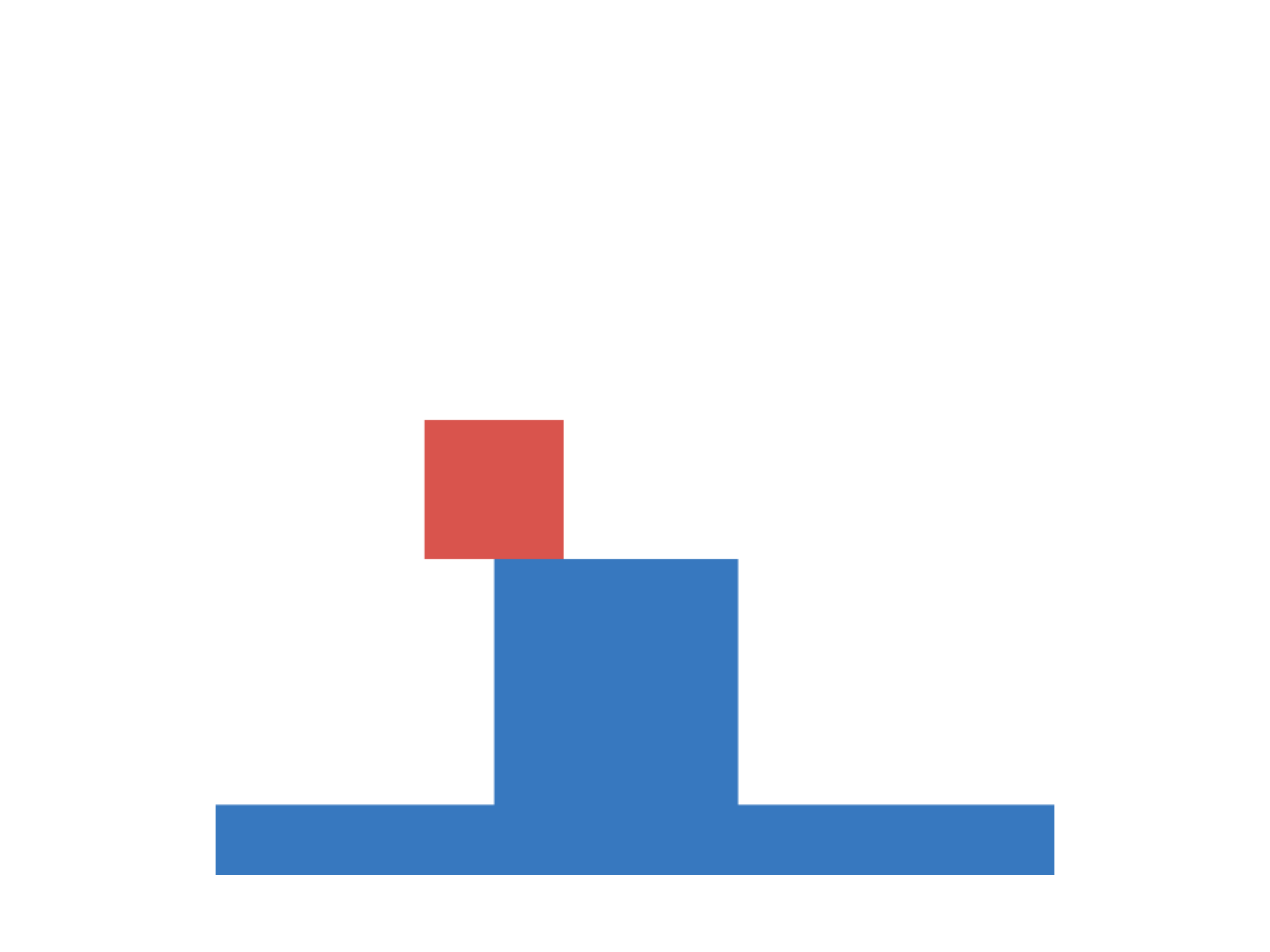} \includegraphics[width=0.39\textwidth]{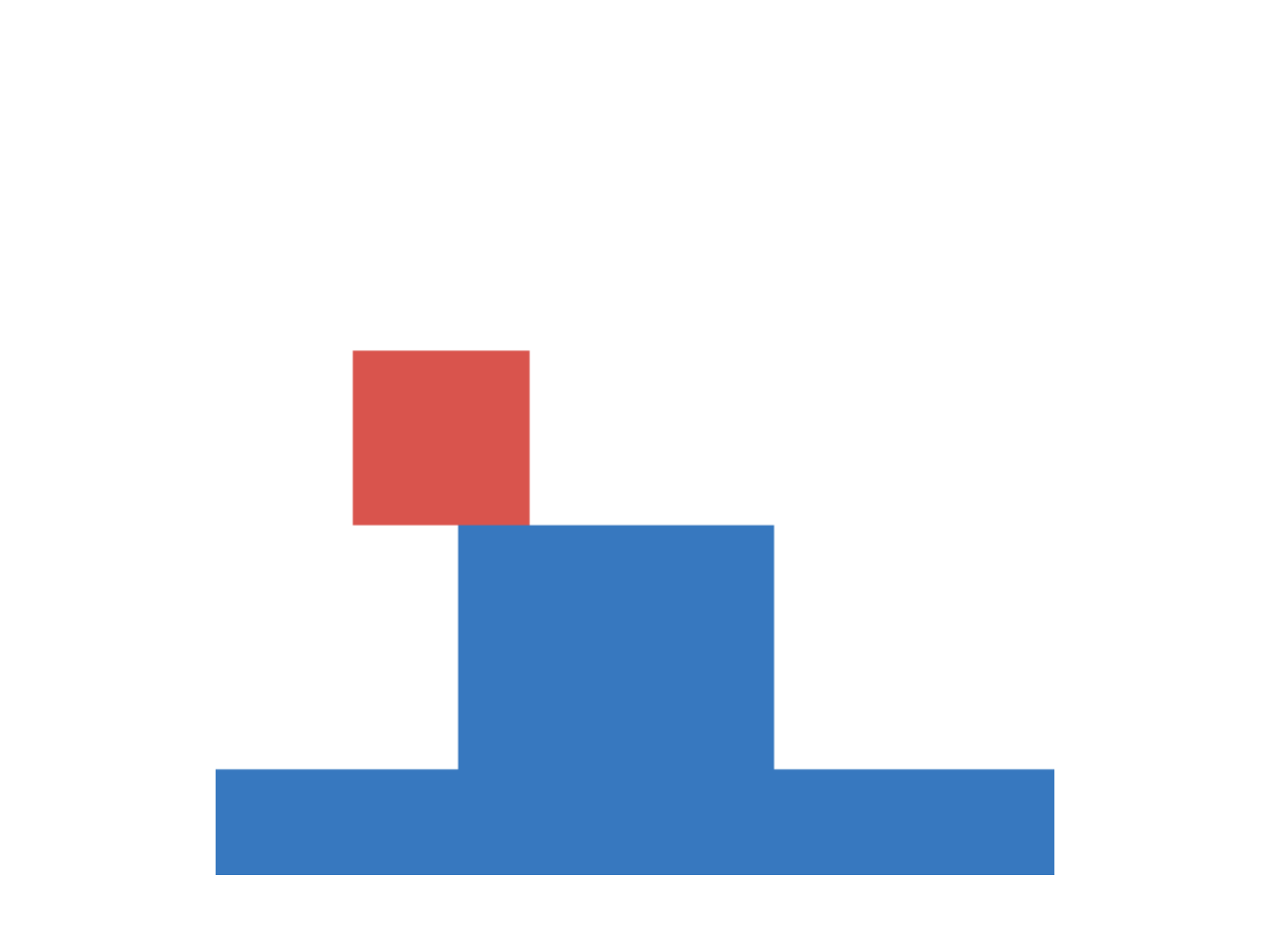} \\
    \includegraphics[width=0.39\textwidth]{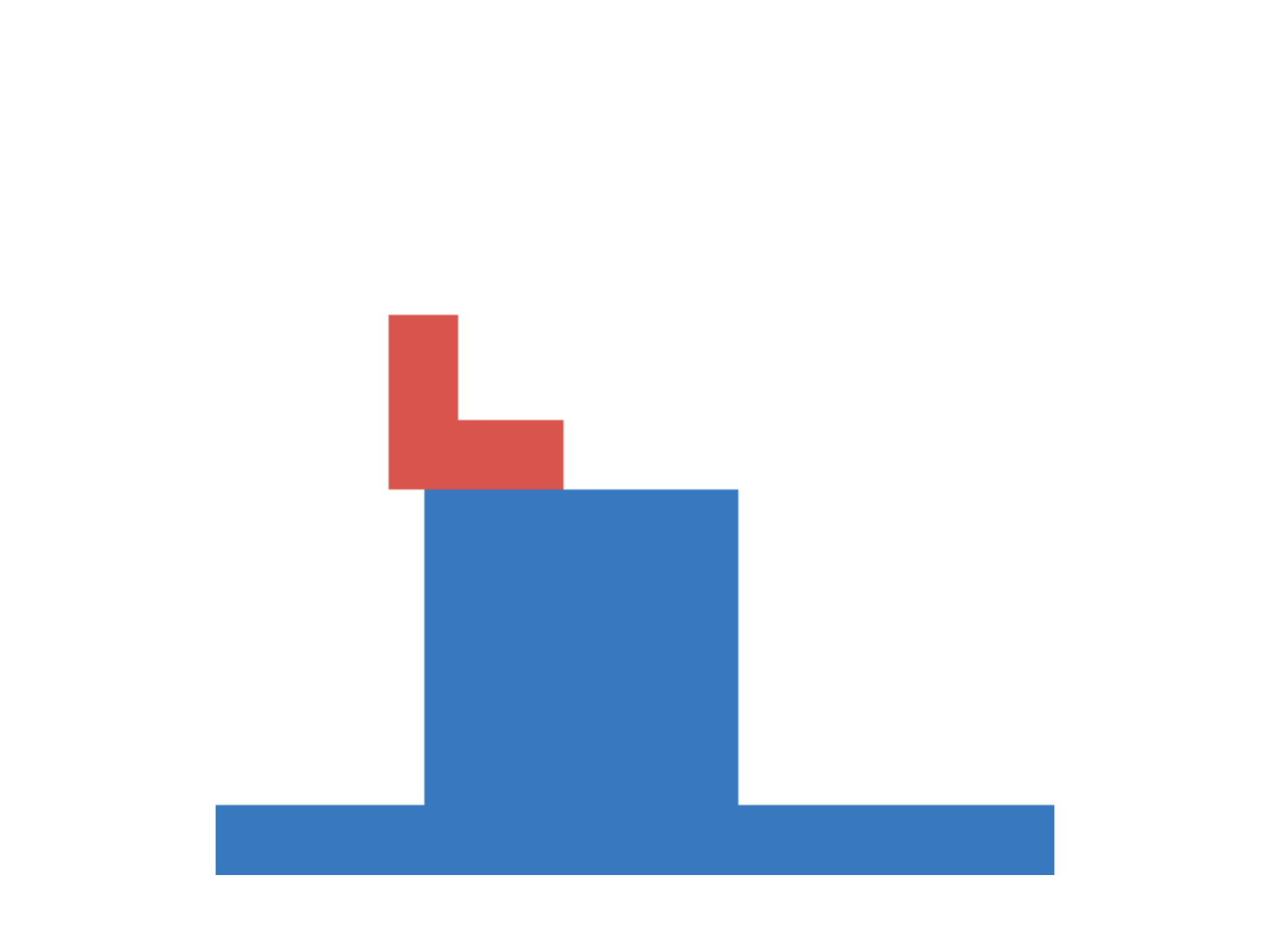} \includegraphics[width=0.39\textwidth]{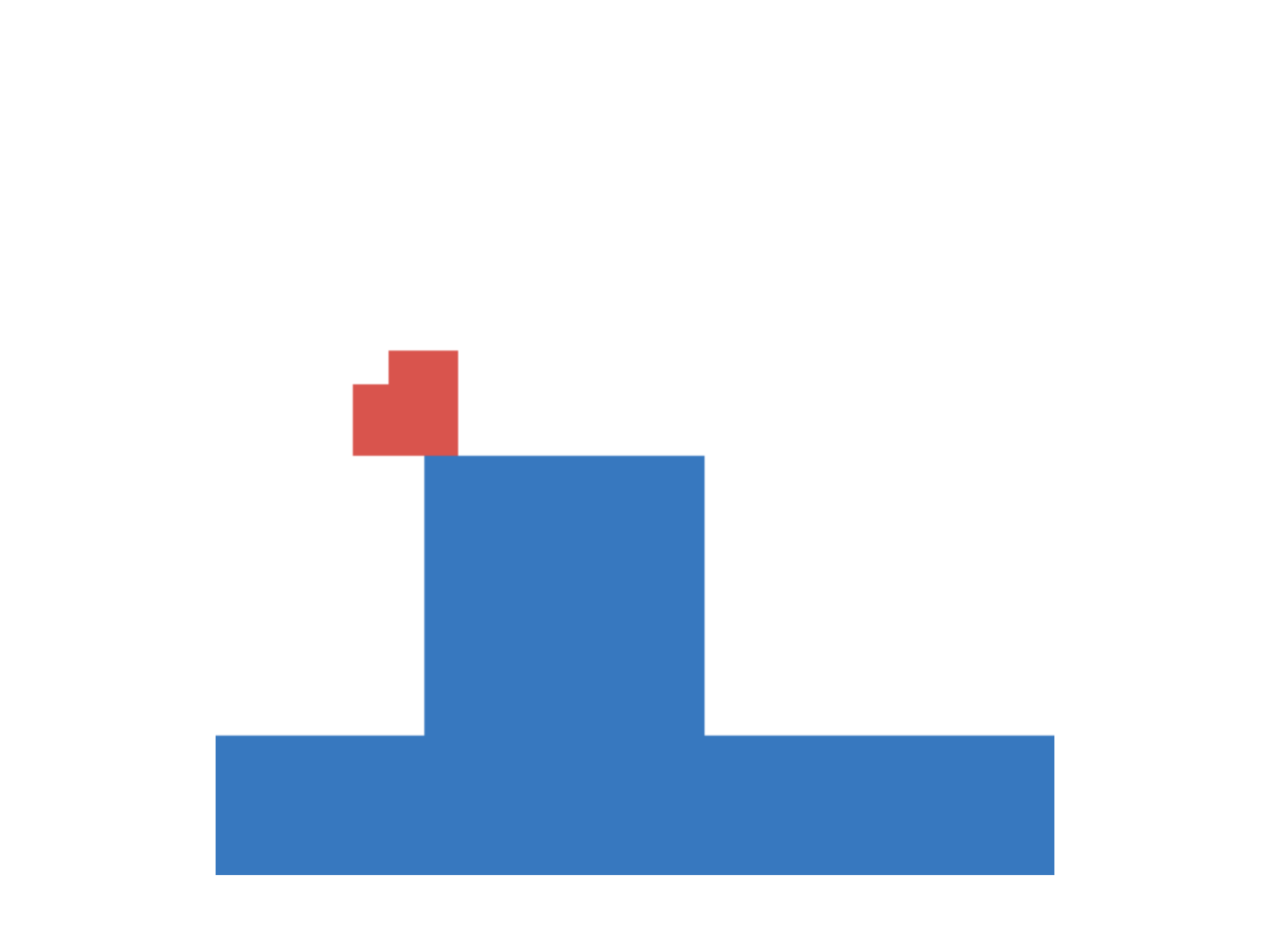} 
    \end{minipage}
    \hspace*{-0.5cm}
    \begin{minipage}{.05\textwidth}
    \small
    \centering
    \vspace{-0cm}
    top \\ \vspace{0.95cm}
    side \\  \vspace{0.95cm}
    amount \\  \vspace{0.95cm}
    prop.
    \end{minipage}
    \begin{minipage}{.43\textwidth}
    \centering
    \includegraphics[width=1\textwidth]{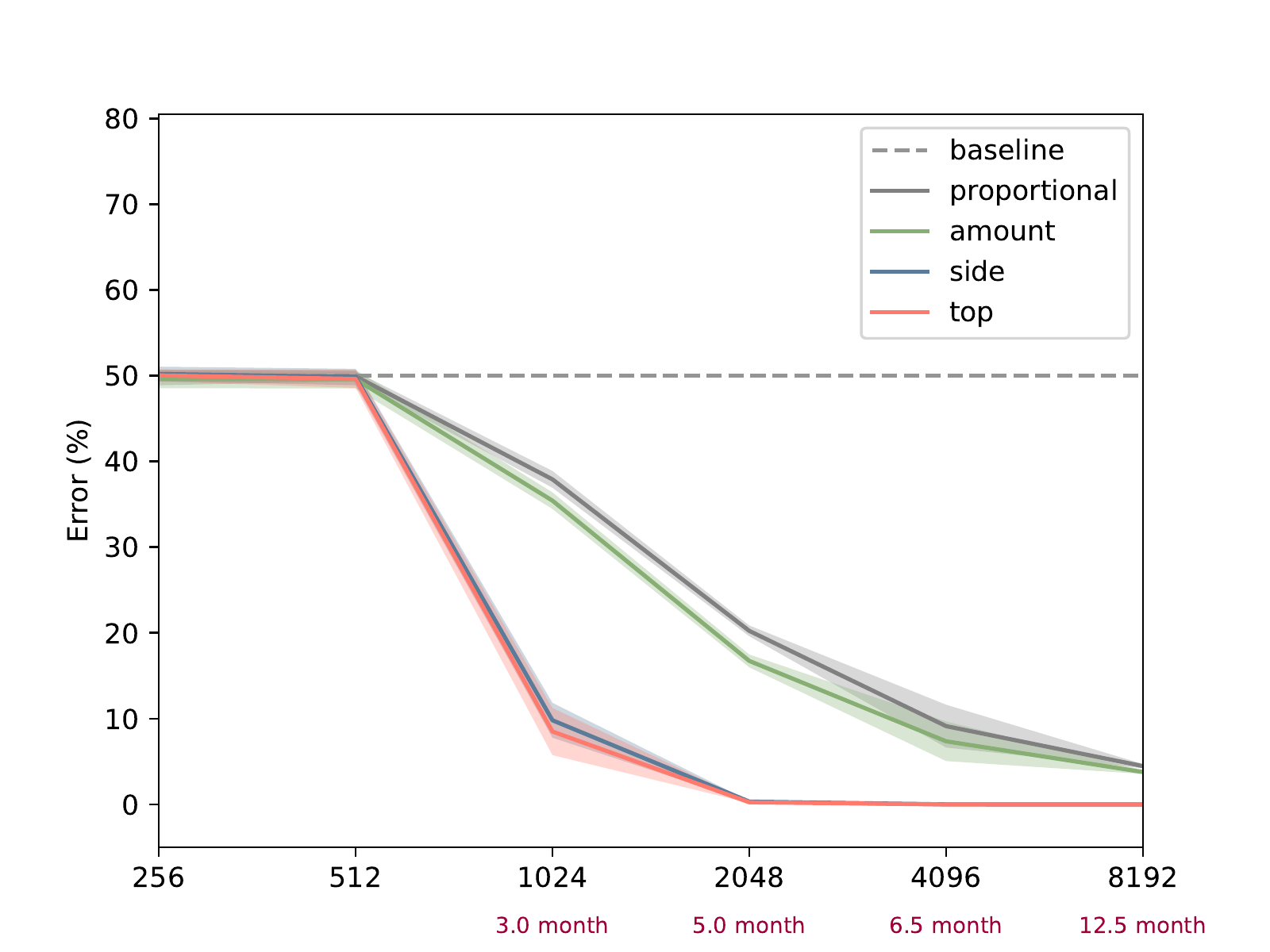}
    \end{minipage}
    
    \caption{Visualization of the support event experiment. \textbf{Left:} Schematic illustration of the original setup. Figure adopted from \cite{baillargeon2002acquisition}. \textbf{Center:} Examples from the artificial data-set alongside the name of the corresponding condition. The left column shows stable configurations, while those in the right column are unstable. \textbf{Right:} Result from training models with access to different amounts of data. Baseline indicates the performance for guessing randomly. Models discover solutions to the easier condition (two upper rows in the left figure) first, and solutions to the harder conditions (two bottom rows in the left figure) later. Note the slight increase in number of samples required for learning between the third and forth condition.
    }
    \label{fig:support}
\end{figure*}

We construct an artificial version of this task as follows. Each input $\mathbf{x}$ corresponds to a $24 \times 24$ image with three channels, containing segregated information about the floor, the cylinder and the screen respectively. Input images show an initial configuration of the scene, in which the cylinder is located on the left side of the screen. Each target $y$ indicates the visibility of the cylinder when passing behind the screen, as it is moved towards the right side of the image. Cylinder height and position, screen position and size and floor level are determined randomly. We generate a training and test set, each consisting of 10000 data-points. Figure \ref{fig:occu} (center) shows examples for each of the conditions. Both sets include 2000 images for each of the three condition, as well as 4000 baseline images (nothing of the screen is removed). This is to ensure, that both $y = 0$ (cylinder is visible) and $y = 1$ (cylinder is not visible) are represented in equal fractions, i.e. the chance of guessing correctly is 50 percent. \\

We train otherwise identical models for $N \in \{256, 512, 1024, 2048, 8192 \}$ until convergence and report results averaged over ten random seeds. The resulting performances on the artificial data-set are visualized in Figure \ref{fig:occu} (right). We observe, that the percentage of incorrect predictions, similar to the developmental progress in infants, decreases first in the easier conditions. The network is able to predict visibility of the cylinder reliably (with less than ten percent errors), if the entire middle section of the screen is removed, for $N$ larger than 512. For $N$ larger than 4096 it is additionally able to predict the correct targets, if the bottom part is removed (corresponding to knowledge of a three months old). This extends to the condition, in which the top part is removed for $N = 8192$ (corresponding to knowledge of a 3.5 months old). Hence we conclude, that in this task the family of BNNs recovers the order of developmental stages observed in infants.

\subsection{Support Events}

Next we take a closer look at infants' knowledge of support events, for which we adopt another experiment of \citeA{baillargeon2002acquisition}. In this task a scene consists of a box and a platform. The box is presented in different positions relative to the platform and the experimenter measures (via violation-of-expectation methods), if infants are able to predict, whether the given configuration is stable or not. Four different conditions are investigated. In the first the box is positioned either on top of the platform or some distance away from it. This condition requires a simple contact/no-contact distinction to make reliable predictions about the stability of the configuration. The second condition involves a distinction between different types of contact. Here the box connects with the platform either on the top (as before) or on the side. The third condition requires judgments based on the amount of contact, i.e. the box is only partially positioned on the platform. The final condition adds an additional layer of complexity, as it involves reasoning about non-rectangular shapes. The different conditions are summarized in Figure \ref{fig:support} (left). \\

According to \citeA{baillargeon2002acquisition}, from three months onward infants knowledge about the stability of a configuration is captured through a contact/no-contact distinction. At this stage they expect the box to be stable if and only if it touches the platform in some way. This initial hypothesis is than refined, as they grow older. At the age of around five months infants begin to distinguish between different types of contacts. They realize, that the box will only be stable, if it positioned above the platform, but not if it touches it on the side. Starting with an age of 6.5 months they are able to take into account the center of mass of simple objects (rectangular boxes) when reasoning about stability. This is extended to more complex, asymmetrical shapes at an age of roughly twelve months. As in the occlusion task, infants start with an initially simple hypothesis of how the laws of physics work, which is subsequently refined to better fit the observed data. \\

In our artificial version of this task inputs $\mathbf{x}$ are represented as $24 \times 24$ images with three channels (one for platform, box and floor) and targets $y$ are a binary indicator of the stability of the given configuration. Floor level as well as the size and position of the platform and the box are randomized. Again we generate a training set of 10000 samples and an equally large test set. In both sets 2500 data-points belong to each of the described conditions. Within each condition the amount of stable and unstable configurations is balanced, leading to a chance of 50 percent for guessing correctly. Example configurations are shown in Figure \ref{fig:support} (center). \\

\begin{figure*}[t]
    \centering
    \begin{minipage}{.26\textwidth}
    \centering
    \textbf{Original Task}
    \end{minipage}
    \begin{minipage}{.39\textwidth}
    \centering
    \textbf{Artificial Data}
    \end{minipage}
    \begin{minipage}{.31\textwidth}
    \centering
    \textbf{Results}
    \end{minipage} \\ 
    \begin{minipage}{.26\textwidth}
    \centering
    \includegraphics[width=0.95\textwidth]{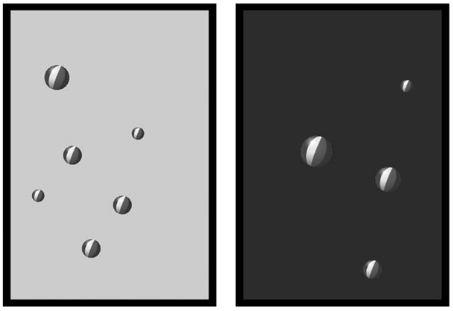}
    \end{minipage}
    \begin{minipage}{.39\textwidth}
    \centering
    \includegraphics[width=0.95\textwidth]{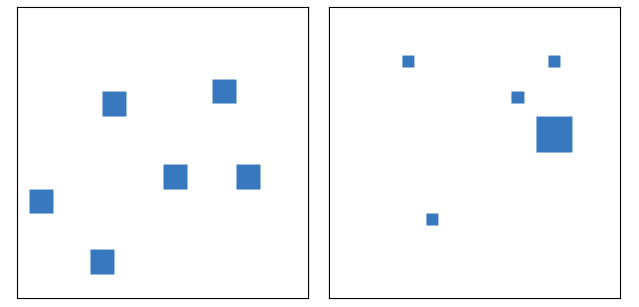}   
    \end{minipage}
    \begin{minipage}{.31\textwidth}
    \centering
    \includegraphics[width=0.95\textwidth]{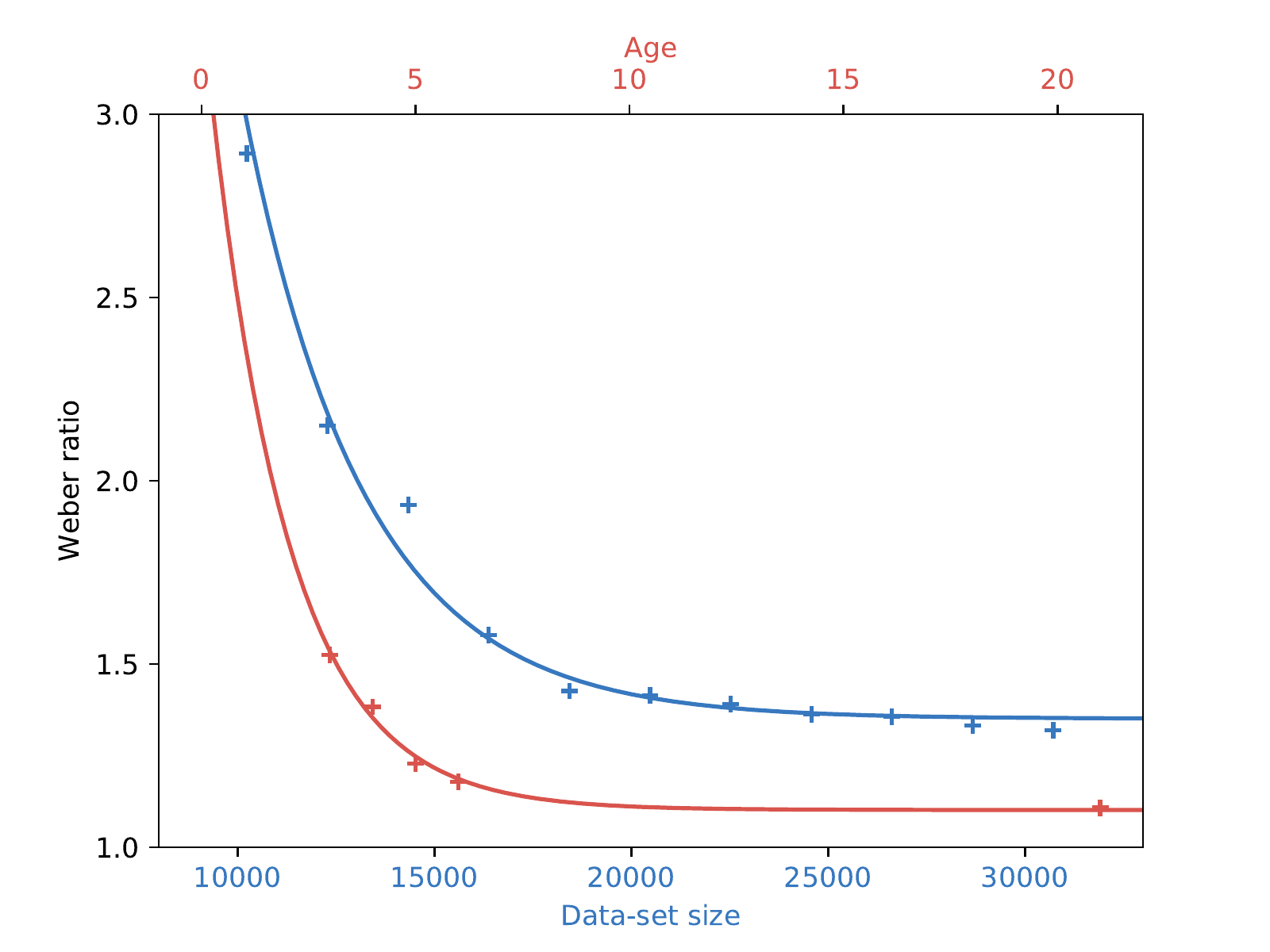}
    \end{minipage}
    
    \caption{Visualization of the approximate number system experiment. \textbf{Left:} Schematic illustration of the original setup. Figure adopted from \cite{halberda2008developmental}. \textbf{Center:} Examples from the artificial data-set. This pair corresponds to a ratio of 6:5. \textbf{Right:} Weber ratios of the experimental data. Red coloring corresponds to estimates from human participants, while blue coloring corresponds to estimates from optimized BNNs. The progression of Weber ratios follows a Weber-Fechner Law and is well described through exponential functions in both cases.}
    \label{fig:ans}
\end{figure*}

As before we train otherwise identical models for $N \in \{256, 512, 1024, 2048, 8192 \}$ until convergence and report result averaged over ten random seeds. Inspecting Figure \ref{fig:support} (right) we observe, that the family of BNNs first discovers solutions to the easier conditions (those where the box is positioned fully on the platform or on the side of it). If we increase $N$ to $4096$ or more, they are also able to reason reliably about stability in both of the center of mass conditions. Note, that the error rate decreases slower, when being exposed to the more complex, L-shaped objects, although this difference is only marginal. We conclude, that the models show pattern similar to the developmental progress of infants, as we increase the data-set size, akin to the observation from experiment 1, although not as pronounced. 

\subsection{Approximate Number System}

Moving away from probing knowledge about physical laws, we next inspect a different domain: children's intuitive counting abilities \cite{halberda2008developmental}. A single trial consists of two images, each containing between 1 and 10 items, and the goal is to determine quickly, i.e. without to much deliberation time, which of the two images contains the larger quantity of items. The display time is adjusted depending on age, such that it is short enough to prevent serial counting. Objects within a trial are identical, but are selected from a set of different objects across trials. Two conditions either control for average item size or the summed continuous area. An example trial from the original task is shown in Figure \ref{fig:ans} (left). \\

Experimental data for this task has been obtained for three to six years olds, as well as for adults \cite{halberda2008developmental}. Human perception is sensitive to the ratio between amounts of objects in the two images and not their difference, i.e. it follows a Weber-Fechner law. Levels of accuracy are measured for different ratios of objects, ranging from 1.11 (10:9) to 2.0 (2:1), from which Weber ratios are estimated. The Weber ratio is the smallest ratio, where a participant is able to identify the correct stimulus in more than 75 percent of the cases. Empirical results indicate, that the Weber ratio decreases during childhood and our ability to distinguish similar stimuli improves over time. Prior work \cite{halberda2008developmental} estimates Weber ratios of around 1.53 in three years olds, which improves to 1.38 in four years olds, to 1.23 in five years olds, to 1.18 in six years olds and to 1.11 in adults. Overall the decrease of Weber ratios is well described through an exponential function of age (see Figure \ref{fig:ans}, right). \\

We created an artificial version of this task, with inputs $\mathbf{x}$ corresponding to two $24 \times 24$ images. For simplicity images contain only rectangular objects and the difference in quantity within a pair of images is always one. Targets $y$ are the number of objects in each images and the final prediction is obtained by comparing expected values between the two estimates. Object sizes are randomly drawn from $\{1, 2, 3\}$ and their position is randomized, such that neighbouring objects do not overlap. We controlled for an equal expected, total area between both images as done in the second condition of the original task. Both training and test set contain 6000 samples for each ratio from $\{$10:9, 9:8, 8:7, 7:6, 6:5, 5:4, 4:3, 3:2, 2:1$ \}$. Examples are provided in Figure \ref{fig:ans} (center). \\

We train one model for each value from $N \in \{4, \ldots , 15 \} \cdot 2048$. Hidden layers have 512 units in this experiment and we found it helpful to initialize weights for all models from a pretrained network. For the resulting models we calculate Weber ratios after estimating accuracy levels of 75 percent via linear interpolation and visualize the results in Figure \ref{fig:ans} (right). We observe an improvement of Weber ratios as the data-set size $N$ increases and conclude, that our models also follow a Weber-Fechner law. The resulting progression is well described through an exponential function\footnote{$y = a e^{-bx + c} + d$, with $x$ corresponding to data-set size or age.} of data-set size, a characteristic shared with the curve obtained from human participants. There is however a small gap between the overall model performance in the artificial task and that of human participants in the original one, even for large data-sets.

\section{Discussion}

We investigated the progress of Bayesian Neural Networks with access to increasingly large data-sets on three different tasks. In all three examples we find an at least partial agreement between the development of our artificial learning systems and findings from the developmental psychology literature. However we also observe some considerable differences. The best performing BNNs in the quantity comparison task, for example, do not reach the level of human adults. We attribute this effect to difficulties for standard neural networks architectures on relational reasoning tasks and hypothesize that recent advances in visual relational reasoning \cite{santoro2017simple} could close this gap. In general we interpret our results as additional evidence for Bayesian theories of cognition and learning \cite{griffiths2008bayesian}. \\

The discriminative models employed in this work require large amounts of input-target pairs to obtain the desired result. Children on the other hand have to operate in a much more data-efficient manner, as they do not have constant access to a teacher providing correct targets. One approach to resolve the question of sources, that children use for learning, are generative models, which are able to discover underlying structures without explicitly provided targets. Whether our results transfer to generative models remains to be seen. Indeed applying generative models in this context would enable us to measure performance in artificial systems directly via  violation-of-expectation methods, as was done with infants in two of our examples. In the future it would be natural to extend our work to more realistic settings and apply different architectures, such as recurrent or convolutional networks. \\

We believe there are exciting opportunities for research on the intersection of machine learning and developmental psychology. On one hand insights from developmental psychology can provide guidelines of how to build more intelligent, human-like systems. The machine learning framework on the other hand enables researchers to formulate normative theories, that can be empirically verified. We already see some progress in these areas. Examples include the usage of violation-of-expectation methods for probing the knowledge of deep networks \cite{DBLP:journals/corr/abs-1804-01128} or the proposal to select training curricula for machine learning systems, based on how children obtain samples \cite{10.3389/fpsyg.2017.02124}.

\section{Acknowledgments}

This work was supported by the DFG GRK-RTG 2271 'Breaking Expectations'.

\bibliographystyle{apacite}

\setlength{\bibleftmargin}{.125in}
\setlength{\bibindent}{-\bibleftmargin}

\bibliography{cogsci_template}

\end{document}